%% file: main.tex
\documentclass[letterpaper,10pt,conference]{resources/ieeeconf}

 \IEEEoverridecommandlockouts                              
\usepackage{amsmath,amssymb,theorem}
\usepackage{graphicx}
\usepackage{algorithm,algpseudocodex}
\usepackage{psfrag}
\usepackage{mathtools}

\usepackage{color,xspace}

\usepackage{siunitx}

\usepackage{caption}
\usepackage{subcaption}

\usepackage[
    backend=bibtex,
    style=ieee,
    sorting=none
]{biblatex}
\makeatletter
\let\NAT@parse\undefined
\makeatother
\usepackage{hyperref}

\input{macros.tex}
\input{resources/common.tex}

\begin{document}
    \title{ReLoki: Infrastructure-free Distributed Relative Localization \\ using On-board UWB Antenna Arrays}

    \author{Joseph Prince Mathew \qquad Cameron Nowzari\thanks{The authors are with the Electrical and Computer Engineering Department, George Mason University, Fairfax, VA 22030, \texttt{\{jprincem, cnowzari\}@gmu.edu. }}}

    \maketitle

    \begin{abstract}
        Coordination of multi-robot systems require some form of localization between agents, but most methods today rely on some external infrastructure. Ultra Wide Band (UWB) sensing has gained popularity in relative localization applications, and we see many implementations that use cooperative agents augmenting UWB range measurements with other sensing modalities (e.g., ViO, IMU, VSLAM) for infrastructure-free relative localization. A lesser researched option is using Angle of Arrival (AoA) readings obtained from UWB Antenna pairs to perform relative localization. In this paper we present a UWB platform called ReLoki that can be used for ranging and AoA-based relative localization in~3D. ReLoki enables any message sent from a transmitting agent to be localized by using a Regular Tetrahedral Antenna Array (RTA). As a full scale proof of concept, we deploy ReLoki on a 3-robot system and compare its performance in terms of accuracy and speed with prior methods.
    \end{abstract}

    \section{Introduction}
    \label{se:Introduction}

    Figuring out where robots are relative to each other is a cornerstone task for coordinating teams of robots. When GPS is not available and no infrastructure or landmarks to rely on for global positioning, the robots must rely on relative localization to support longer missions such as search-and-rescue or environmental monitoring. A priori information about the environment may also not be available. 
    This paper proposes and validates a novel method using only on-board Ultra-Wide Band (UWB) antenna arrays to relatively localize robots in a 3D environment without obstructions.

    Our interest is in a lightweight real-time distributed and completely on-board sensing solution to 3D relative localization that can plug-and-play into any existing multi-robot platform. Acquiring distance estimates based on Time of Flight (ToF) differences in UWB sensors is a commonly used method today~\cite{Chen2022,Guler2021, Li2020, Barral2019}. A lesser explored area of interest for UWB systems is using multiple antenna arrays to leverage its Phase Difference of Arrival (PDoA) \cite{Dotlic2018}. 
    As far as the authors are aware, this paper is the first to implement an infrastructure-free full 3D relative localization method on real hardware using only UWB sensors. We propose ReLoki as a novel distributed relative localization system that is implemented using UWB modules capable of obtaining PDoA estimates when communicating. In~\cite{Luo2022} the authors propose a theoretical method in which an antenna array commonly used in acoustic localization called the \textbf{Regular Tetrahedral Antenna Array (RTA)}~\cite{Gregory1997, hioka2004, Datta2010, Freire2011} can be leveraged to obtain 3D relative position estimates. We note that the algorithm can be modified to be used with any 4-antenna array configuration but the geometry also has an affect on performance. This paper will compare the RTA array with a previously developed orthogonal antenna array~\cite{PrinceMathew2023}.

    \begin{figure}[t]
        \centering
        \includegraphics[width=0.8\linewidth]{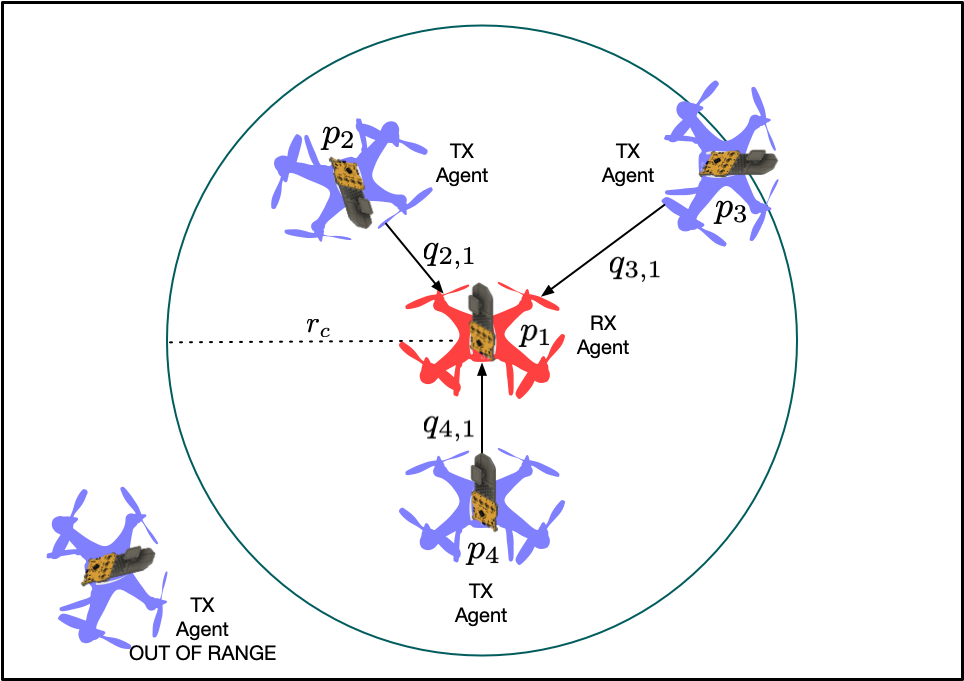}
        \caption{Illustration of the relative localization problem with ReLoki attached to any existing motion platform. Here the RX agent senses the relative positions $\relpos$ of the TX agents w.r.t its body frame whenever a message is received from~$j$.}
        \label{fig:UWBScenario}
    \end{figure}
   
    \subsection{Related Works}
    A common method of obtaining range measurements using UWB systems is the Tag-Anchor model \cite{Guler2021, Li2020, Zhao2021}. Anchor nodes are set up at predefined locations and the Time Differences of Arrival (TDoAs) are used to estimate the relative position relative of these tags. Another implementation called Concurrent AoA estimation uses a single RX antenna on the robot as a tag and a multi-TX antenna array on the anchors situated at known locations in a room~\cite{Heydariaan2020,Smaoui2021} with small delays between transmissions. 
    Unfortunately, all the above methods require pre-existing infrastructure, making them unsuitable for our motivating applications.

    To remove the need for infrastructure, we see a multitude of sensor fusion methods~\cite{Nguyen2021, Martel2019, Li2020, Cossette2021, Cao2021}, where the ranging measurement obtained from a single UWB TX/RX system on each robot is combined with some form of SLAM using IMU and/or visual data. These systems have decent performance; however, they are far from plug-and-play and require very complex computation requirements. Additionally, these systems require multiple readings from different points in the environment to estimate position which contributes to an initial windup time before these sensors get lower errors in their positional estimates.

    There have also been multi UWB range measurement systems aimed at estimating the full 2D position (with pose) \cite{Fishberg2022} and 3D position \cite{Phalak2020, Kohlbacher2018, Xianjia2021} of robots. These systems use UWB antennas placed at pre-defined locations of the robot i.e. the body of the robot is part of the antenna array structure. The difference in the ToA measurements on these different antennas corresponds to the Angle of Arrival of the source. The bearing angle errors in this case decrease with higher separation between antennas in a the array. This leads to large and bulky systems that needs bigger motion platforms for optimal operation. For small sizes, the bearing estimation based on these ToA measurements has much higher errors than corresponding PDoA-based systems as shown in \cite{Dotlic2018}. 

    Finally, there is the PDoA-based Angle of Arrival coupled with ToA range measurement that is used by our proposed ReLoki. The idea is to use phase differences measured by antennas separated by a distance of roughly half the carrier wavelength to estimate the AoA of the signal. An azimuth-only AoA version is considered in~\cite{Dotlic2018}. Here we see a one-to-two mapping of the measured phase difference in the range $[-\pi, \pi]$ to actual azimuth angle to in the range $[-\pi, \pi]$ which means, for a single phase difference value 2 azimuth estimates exists (azimuth ambiguity). Most similar to our work~\cite{Tiemann2020} is an extension of this implementation which uses one more orthogonal pair (total of 3 antennas) to estimate the elevation angle, but, the same issue of azimuth ambiguity still exists. Our previous work~\cite{PrinceMathew2023} extended this to three antenna pairs (total of 4 antennas) orthogonal to each other which enables 3D estimates but had high errors at higher elevation angles. The use of the RTA Array introduces more antenna pairs with the half carrier wavelength separation that can be effectively used and we expect to have better performance. When dealing with the antenna separation for the RTA Array, existing work on warped PDoA measurements for antennas with spacing greater than half carrier wavelength~\cite{Luo2022} has shown theoretical performance gain over simple PDoA based estimation without wrapping (for antenna spacing less than half carrier wavelength), but this comes at the cost of using a computationally expensive non-convex search strategy to address phase ambiguity resolution. As a first proof of concept, we instead implement a simpler approach with antenna spacing less than half carrier wavelength to facilitate the low weight of the hardware and faster estimation.

 \subsection{Contributions}
    The main contributions of the paper are as follows.
    \begin{itemize}
        \item First proof of concept in hardware (weight: 65g) of an RTA array on UWB system that combines Two-Way-Ranging (TWR) and PDoA to estimate relative position.
        \item Analyze the performance of the system and compare with existing Orthogonal antenna array system~\cite{PrinceMathew2023}. 
        \item Full scale proof of concept experiments with 2 moving and 1 static agent, achieving an error tolerance of less than~$25$cm for range estimates and $15 \unit{\degree}$ for bearing estimates in the operational range of $(-60\unit{\degree}, 60\unit{\degree})$ elevation.
    \end{itemize}

    \section{Multi Agent Relative Localization via UWB}
    \label{se:MARL_UWB}
    Consider a set of robots $\robots = \{1, 2, .. i, .. \robotsnum\}$ freely moving in a given environment with no obstacles. Let the 3D position of robot~$i \in \robots$ be defined as $\robotposown \in \real^3$. 
Each agent is equipped with an UWB sensor and a 4-antenna array to communicate with each other. 
The goal of each agent is to use this UWB system to get an estimate $\relposest$ of the relative position $\relpos \in \real^3 := \robotbodyframe (p_i - p_j)$ of its neighbors with respect to its own body frame, where $\robotbodyframe$ denotes the rotation from global to the body frame of the robot doing the relative localization sensing. 

    {
        \problem[3D relative position estimate]
        Using the UWB sensing system mentioned above, find a co-designed Two Way Ranging (TWR) + Phase Difference of Arrival (PDoA) pinging scheme and an online estimation algorithm that allows agent $j$ to sense the relative position $\relpos$ to agent $i$, whenever agent $i$ initiates a Ranging request. The combined algorithm and hardware should minimize the relative localization error cost
        \begin{align}
            \label{eqn:objective}
            J_i^{\text{err}} = \sum_{i \in \robots \setminus j} E[||\relpos - \relposest||]. 
        \end{align} 
    }



    

    \section{ReLoki Proof of Concept}
    \label{se:ReLoki_Desc}
    
\begin{figure}
        \centering
        \begin{subfigure}{0.49\linewidth}
            \includegraphics[width=0.9\linewidth]{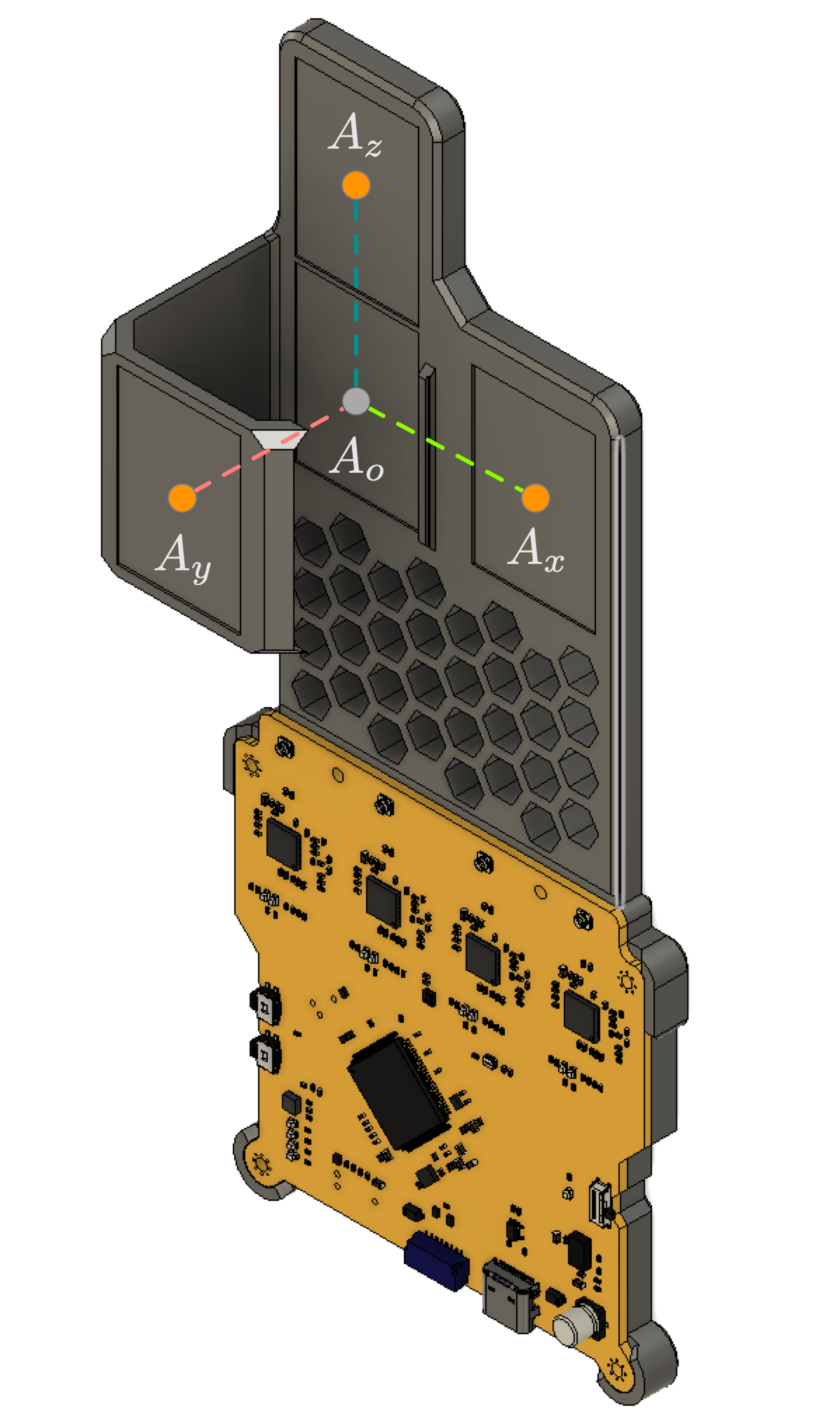}
            \caption{Orthogonal Antenna Array}
            \label{sfig:OrthoAssembly}
        \end{subfigure}
        \begin{subfigure}{0.4\linewidth}
            \includegraphics[width=0.9\linewidth]{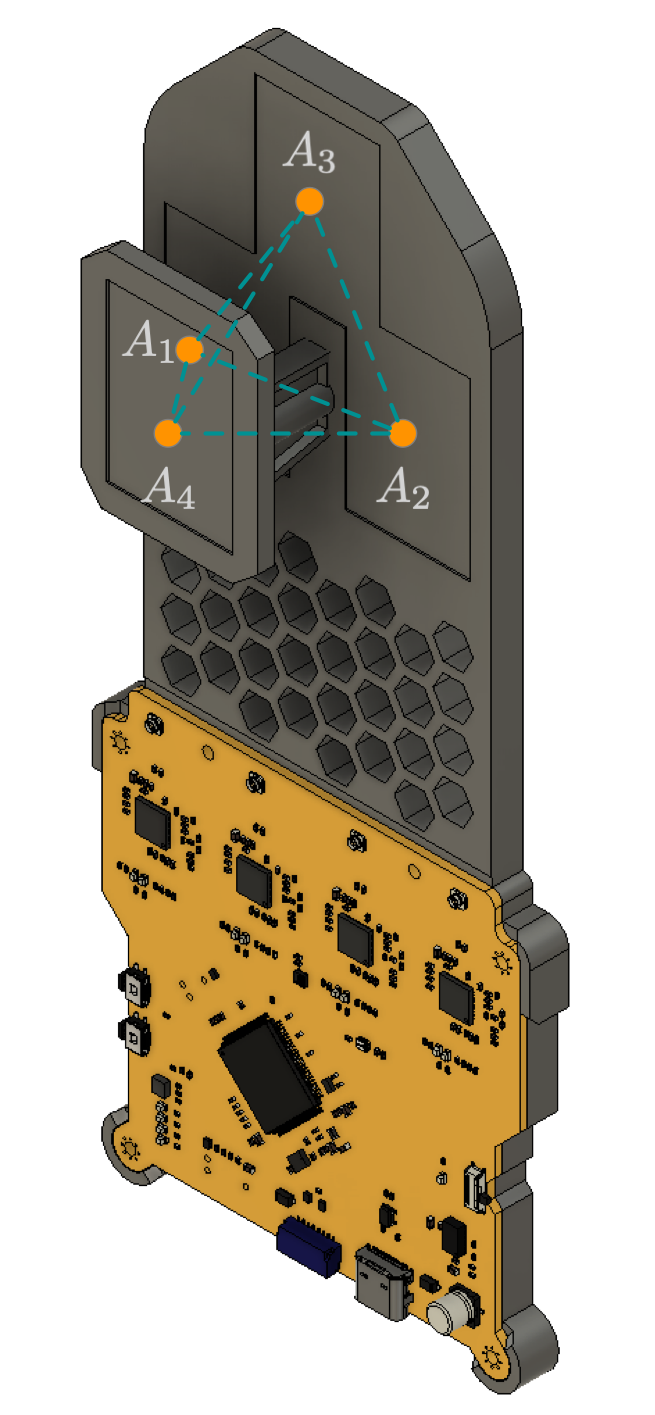}
            \caption{RTA Antenna Array}
            \label{sfig:RTAAssembly}
        \end{subfigure}
        \caption{Examples of two 4-antenna antenna configurations that can be used with ReLoki.
        }
        \label{fig:AntennaStructure}
    \end{figure}

ReLoki uses a set of UWB sensors to determine the range and direction to a source when it receives an incoming communication. Each node in the system consists of two parts: 1) an antenna array that is used to transmit/receive (TX/RX) data, and 2) a base platform consisting of 4 UWB modules, capable of computing the phase of arrival of an incoming signal, connected to a processing subsystem that performs all the computations. 
Here we discuss the RTA, but it should be noted that any 4-element antenna array can be used with this system, which is the minimum number required for full 3D localization~\cite{Luo2022}. An illustration of the RTA antenna array and our previously designed orthogonal antenna array~\cite{PrinceMathew2023} is shown in figure~\ref{fig:AntennaStructure}. 

    \subsection{Relative localization using 4-antenna RTA}
    \label{se:RL_RTA}
    For localization of a source, we need ranging and direction/bearing information. Obtaining range measurements is straightforward. To determine the direction to a source we take advantage of the RTA array for UWB and comparing Phase Difference of Arrivals (PDoA) between different antenna pairs. This can be done each time an agent~$i$ initiates what we call a Relative Position Ping (RPP).
    
    The ToA of a radio signal between two nodes is calculated using an asynchronous Two Way Ranging (TWR) between the nodes which is already well established and known to provide centimeter-level accuracy~\cite{Decawave2015,Malajner2015, Chantaweesomboon2016}.
    
    
    For PDoA measurements, ideally the distance between all antenna pairs are set to half the wavelength of the carrier $\carrierwl / 2$ to avoid wrapping (phase difference estimated goes over $\pi$ or $-\pi$, thereby wrapping to the opposite sign). Additionally, the carrier frequency should be in sync between all UWB transceivers. Assuming we have the phase of arrival~$\uwbphcalc{n}, n \in \{1,2,3,4\}$ for all 4 antennas in the RTA array, we can compute the phase difference for all ${4 \choose 2} = 6$ antenna pairs as with
    
    
    \begin{align}
        \label{eqn:phaseDif}
        \uwbphdiff = \uwbphcalc{n} - \uwbphcalc{o} + \uwbphcal,\quad n \neq o.
    \end{align}
    It may be the case that the phase difference computed for a source normal to the antenna pair is not zero. We use a bias cancellation term $\tilde{\Phi}_j^{n, o}$ to compensate for this. The bias compensation can be found by calibrating each antenna pair by first measuring the phase angle without any bias for all possible angle pairs. These are compared with the true angles they should output based on their geometry and the bias factor~$\uwbphcal$ is found to minimize the least squared error.

    \begin{figure}
        \centering
        \includegraphics[width=0.68\linewidth]{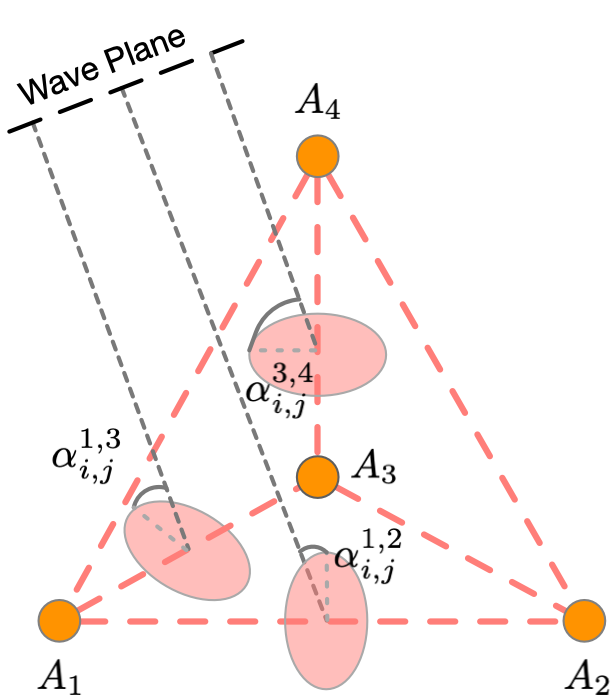}
        \caption{Regular Tetrahedral Array with three 3 sample incidence angles used to estimate the bearing angle.}
        \label{fig:ReLoki_IncidenceAngles}
    \end{figure}

    Subsequently, the angle of incidence for all 6 antenna pairs is computed using \eqref{eqn:CIR_IncidenceAngle}. We have a sample illustration of angle of incidence for 3 antenna pairs shown in figure~\ref{fig:ReLoki_IncidenceAngles}.
        \begin{align}
            \label{eqn:CIR_IncidenceAngle}
            \uwbincang = \frac{1}{0.95}\arcsin\left(\frac{\uwbphdiff}{\pi}\right), \quad n \neq o.
        \end{align}

    \begin{algorithm}[t]
        \caption{Angle-of-Arrival (AoA) Estimation}
        \label{alg:AOA_Extimation}
        \hspace*{\algorithmicindent} \textbf{Input} Phase of arrival at each antenna $\uwbphcalc{n}$ \\
        \hspace*{\algorithmicindent} \textbf{Output} unit vector bearing of neighbor $\ubear$
        \begin{algorithmic}[1]
            \State Calculate the phase difference $\uwbphdiff$ with the bias compensation using~\eqref{eqn:phaseDif}
            \State Calculate the incidence angle $\uwbincang$ at each antenna pair using~\eqref{eqn:CIR_IncidenceAngle}
            \State Get the raw estimate of the unit bearing vector~$\ubearraw$ using \eqref{eqn:RTAIncAng2Bearing}
            \State Normalize the unit vector~$\ubear$ to length $1$ using \eqref{eqn:RTABearingNormal}
            \Return $\ubear$
        \end{algorithmic}
    \end{algorithm}

    To note is the use of a fraction of $\frac{1}{0.95}$. In the actual design of the antenna array, we put the antenna spacing to be $0.95 \carrierwl / 2$. This is done to avoid the calculated phase differences going over $[-\pi, \pi]$ due to the noises in phase calculation algorithm as mentioned in \cite{Dotlic2018}. Using the 6 PDoA measurements and the specific geometry of the antenna array, we can obtain the unit vector $\ubear = [\ux, \uy, \uz]$ pointing towards the source by solving the redundant equations
    
    
     \begin{align}\label{eqn:RTAIncAng2Bearing}
     \begin{bmatrix}
                 1           & 0             & 0 \\
                 (-1/2)      & 0             & (\sqrt{3}/2) \\
                 (-1/2)      & 0             & - (\sqrt{3}/2) \\
                 (1/2)       & (\sqrt{3}/6)  & (\sqrt{6}/3) \\ 
                 (1/2)       & (\sqrt{3}/6)  & - (\sqrt{6}/3) \\
                 0           & (\sqrt{6}/3)  &  (1/\sqrt{3}) \\
             \end{bmatrix}
              \begin{bmatrix}
                 \uxm \\
                 \uym \\
                 \uzm 
             \end{bmatrix}
             = 
             \begin{bmatrix}
                 \sin(\uwbincangname{2}{1}) \\
                 \sin(\uwbincangname{3}{2}) \\
                 \sin(\uwbincangname{1}{3}) \\
                 \sin(\uwbincangname{4}{1}) \\
                 \sin(\uwbincangname{4}{2}) \\
                 \sin(\uwbincangname{4}{3}) 
             \end{bmatrix}
     \end{align}

    In ideal cases without noise, all six equations should point to a unique solution but with noisy measurements this naturally becomes over-constrained. It is noted for UWB modules that the noise is directly correlated with the magnitude $|\uwbincang|$~\cite{Dotlic2018} (i.e., when the magnitude is low the estimate of incident angle has very small error, however, estimates over $70 \unit{\degree}$ show high errors). Hence, for a specific set of phase differences received, we remove from \eqref{eqn:RTAIncAng2Bearing} the output antenna pair row with very high phase difference $|\uwbphdiff| > \uwbphdifflim$. We experimentally find a threshold~$\uwbphdifflim = 165\unit{\degree}$ that works well. Next, we use a pseudo inverse of $A$ based on the remaining pairs with valid phase difference value in the system of equations and approximately solve~\eqref{eqn:RTAIncAng2Bearing} to obtain the normalized direction estimate
        \begin{align}
        \label{eqn:RTABearingNormal}
        \um &= \frac{\umm}{\sqrt{\sum_{s \in {x, y, z}} (\us)^2}}.
    \end{align}

    The full AoA Estimation relative localization process is formalized in algorithm~\ref{alg:AOA_Extimation}. With the range measurement $\urange$ obtained from TWR and the direction $\ubear$ obtained from algorithm~\ref{alg:AOA_Extimation}, we get the estimated relative localization of the system as $\hat{\relpos} = [\urange \ux, \urange \uy, \urange \uz]$.

    \begin{figure*}[t]
        \centering
        \includegraphics[width=\linewidth]{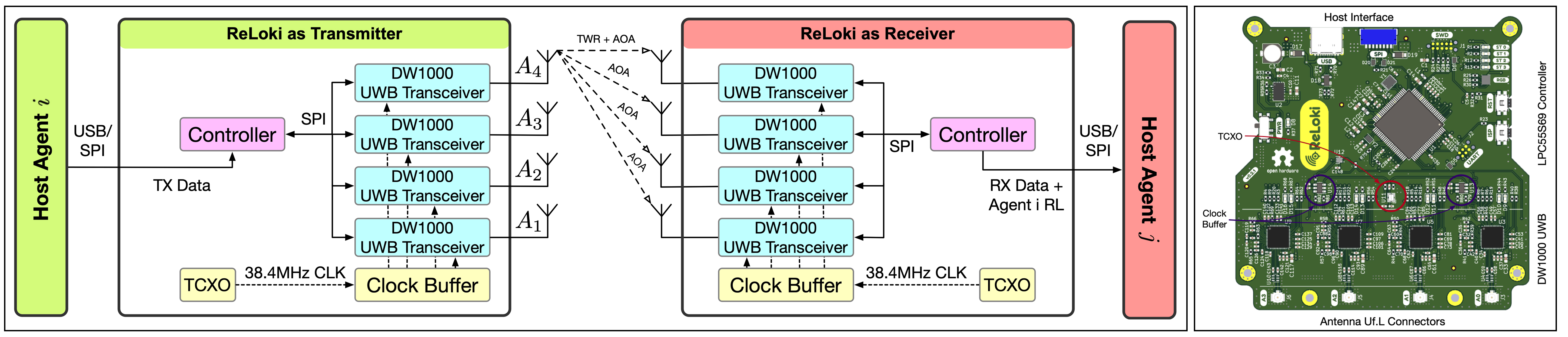}
        \caption{On the left we have ReLoki hardware block diagram showing the components. Here we start with the agent $i$ initiating communication request. ReLoki connected to Agent $i$ transmits the data, ReLoki broadcasts the information to all agents in the neighborhood, the receiving ReLoki decodes the message and the localization data and intimates the receiving agent $j$. On the right we have the ReLoki PCB}
        \label{fig:ReLoki_BlockDiagram}
    \end{figure*}

    \begin{figure}[t]
        \centering
        \includegraphics[width=0.8\linewidth]{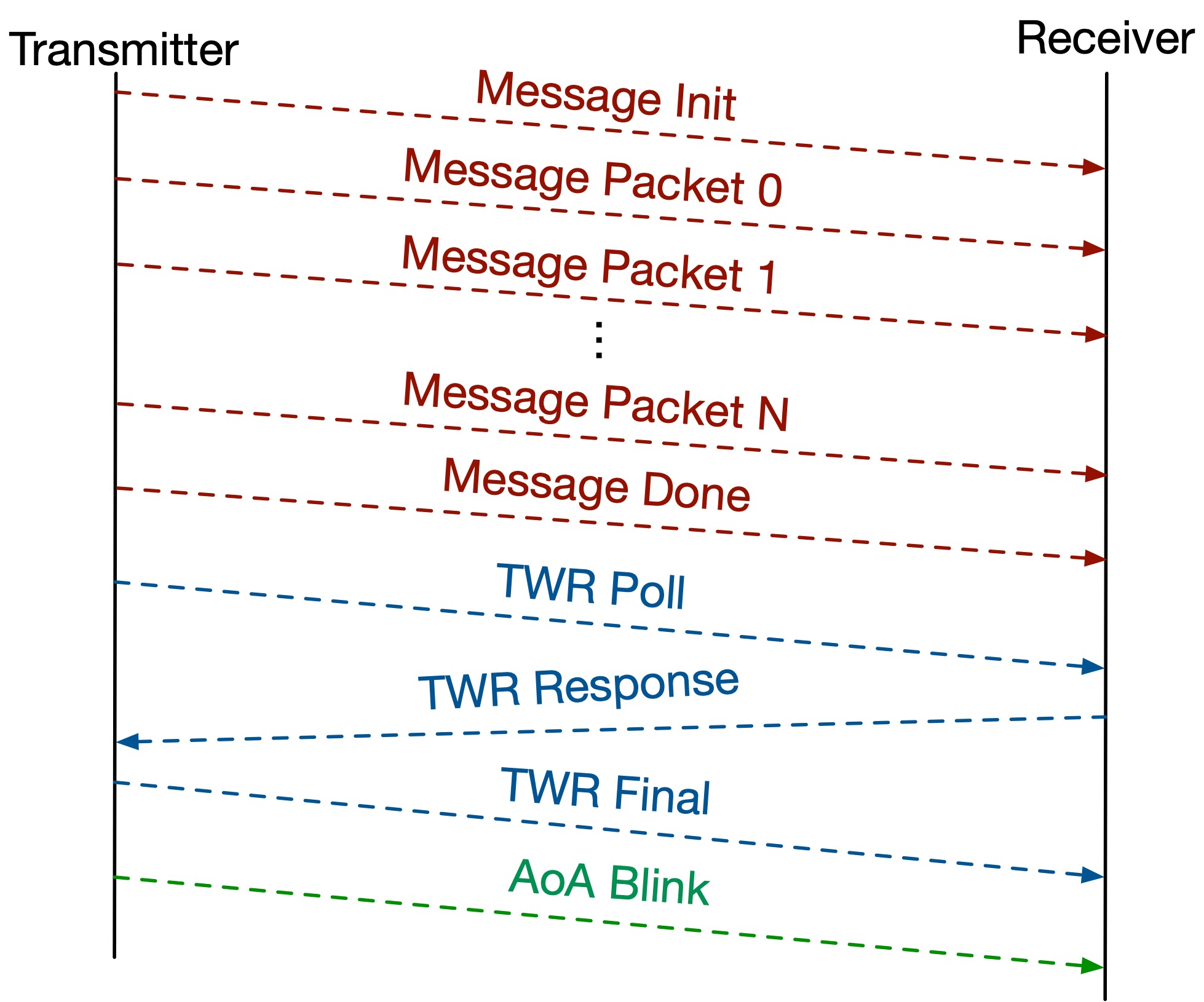}
        \caption{Timing Diagram showing the different phases of transmissions. Message phase is shown in red, ranging phase in blue and bearing phase in green.}
        \label{fig:Reloki_TimingDiagram}
    \end{figure}

    \subsection{Messaging Protocol}
    \label{se:message_protocol}

    Here we discuss our specific messaging protocol ReLoki uses with the RTA. When acting as the transmitter, only one antenna~$A_4$ is used. All transactions begin with a request to initialize a message called message init. Subsequent data transfer consists of three phases: message, ranging, and bearing as shown in figure~\ref{fig:Reloki_TimingDiagram}. In the message passing phase the message to be transmitted is properly formatted to a 802.15.4A packet and sent. In a typical 802.15.4A transmission, the maximum payload is 127 bytes of data. Hence, we split the original message to packets of size 120 bytes and transmit the data. The extra 7 bytes can be used for packet header data and Cyclic Redundancy Check(CRC). The second phase is the ranging phase using the TWR protocol mentioned in section~\ref{se:RL_RTA} and allows the range sensing to the source. For the phases mentioned above, the receiver uses only its~$A_4$ receiver. The third phase is the bearing phase used to measure the bearing of the transmitter from the receiver and here all 4 antennas are used on the receiver side. This entire transaction is called a RPP. Failure in any point of the protocol will lead to termination of transaction.

    \subsection{ReLoki Hardware Platform}
    The above method can be implemented on different hardware, for our proof of concept we use four DW1000 modules per agent, each interfaced with a LPC55S69 micro-controller. All the DW1000 modules are clocked from the same clock source ATX-13-38400 of 38.4MHz through a clock distribution buffer. This is to ensure that the generated carrier wave on all the DW1000s on board are in sync. To allow flexibility of connecting any 4 element antenna array, Uf.L connectors are used on the platform. We are using Taoglas AccuraUWB FXUWB10.07 as antenna elements. All the computations required for setting up the data transfer and the Relative Localization estimation is implemented on this controller platform. We show a block diagram of the system and the PCB in figure~\ref{fig:ReLoki_BlockDiagram}. The total weight of one node is only $65\unit{g}$ including PCB, antennas and the 3D printed antenna holders. 


    In DW1000, when the antenna array receives data on all 4 antennas from a single source, all the UWB modules in the receiver computes the Complex Impulse Response (CIR) at the first path (FP) in the accumulator memory~\cite{Dotlic2018}. We can hence compute the phase of arrival at any antenna $n$ as

    \begin{align}
        \label{eqn:CIR_PhaseCompute}
        \uwbphcalc{n} = \arctan\left(I_n(\uwbfptime) + i . Q_n(\uwbfptime)\right) - \beta_j^n.
    \end{align}

    where, $\uwbfptime$ denotes the time in accumulated memory of DW1000 which it has determined is the first path reception and $I_n(.)$ and $Q_n(.)$ denotes the real and complex magnitudes. Additionally, there is a correction factor $\beta_j^n$ applied at each UWB to the calculated phase called Synchronous Frame Detection (SFD) angle. This phase correction is an artifact of the FP detection in the DW1000 module~\cite{Dotlic2018}.

    When agent $i$ has to send some message to its neighbors, it will send the data to the ReLoki attached via its host interface. The sending ReLoki initiates the transaction if possible using its antenna assigned for sending $A_4$. With the transaction happening as mentioned in section~\ref{se:message_protocol} and the localization happening as mentioned in section~\ref{se:RL_RTA}, the receiving ReLoki will discern the message being sent and the Relative Localization data. It then combines both into a single message for the host and intimates the host connected of the received message. The host can read this data and use it as required in its own control loop.

    \subsection{Muti-agent support}
    The UWB based position sensing can only be performed by one sender-receiver pair at a time. To extend the same to multiple agents, we need to have a form of time multiplexing available to the agents. All agents who are expected to receive data will wait for a Message init packet. If a valid message init was received, it proceeds to data reception. Hence, the specific states the controller is in will dictate if there is an ongoing signal. Thus, for any ReLoki wanting to transfer a message, if that ReLoki is already participating as a receiving agent, then the transmission will be delayed for a random time and then this check is performed again. The transmission happens only if the airway is clear with this scheme. This is a specific implementation of carrier sensing based on the UWB states that is supported by DW1000 and we leverage this here for Carrise Sense Multiple Access(CSMA).
    
    \section{Performance Analysis and Experiments}
    We test the performance of the ReLoki on two fronts. First, we experimentally measure the errors in estimation and generate a covariance map. Second, we check the max communication frequency possible for the system.
    
    \begin{figure}[h]
        \centering
        \includegraphics[width=0.9\linewidth]{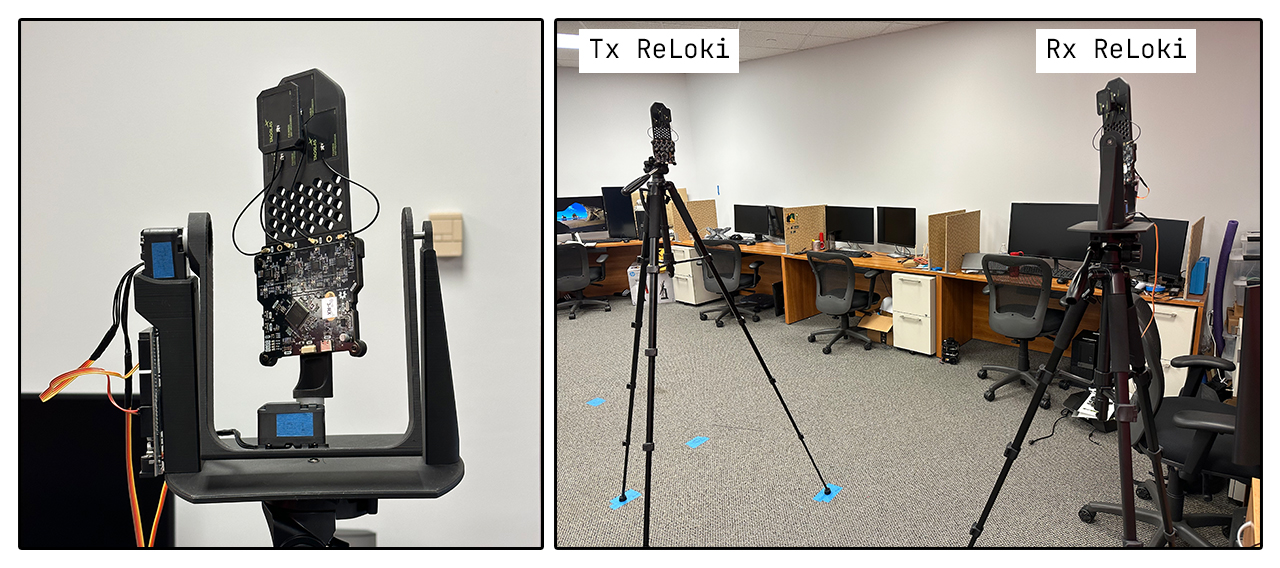}
        \caption{ReLoki Experimental Setup for Covariance measurement. On the left we have the pan and tilt mechanism and on the right we have the test setup for $1.5\unit{m}$ range from source.}
        \label{fig:ReLoki_ExpmSetup}
    \end{figure}

The covariance map for ReLoki is the map of the expected error at different regions of the sensing domain. Having a lower value in this covariance map will translate to a smaller cost defined by~\eqref{eqn:objective}. To determine the covariance map, we set up an experiment where one ReLoki is mounted in a pan-tilt setup as shown in figure~\ref{fig:ReLoki_ExpmSetup}, which allows to test all possible orientations the source will have with respect to the receiving agent. We take 50 readings with pan range $[-180\unit{\degree}, 180\unit{\degree}]$ and elevation readings $[-90\unit{\degree}, 90\unit{\degree}]$ in steps of $15\unit{\degree}$ and ranges from $1.5\unit{m}$ to $7.5\unit{m}$ in steps of $1\unit{m}$. 

    \begin{figure*}[t]
        \centering
        \includegraphics[width=0.9\linewidth]{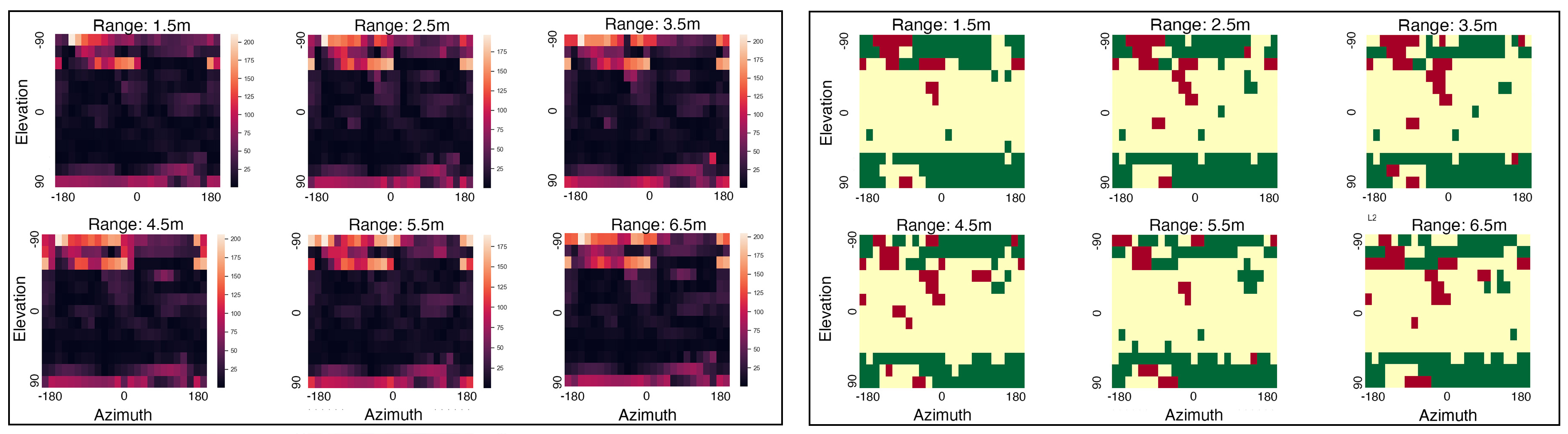}
        \caption{On the lefe we have RTA Antenna Covariance Map. A darker color means lower error. On the right we show the comparison of RTA Antenna array to Orthogonal Antenna array. Here green boxes represent lower errors for RTA than that in orthogonal array and red represents the opposite. Yellow represents comparable performance (combined azimuth and elevation difference within $10\unit{\degree}$) between both.}
        \label{fig:ReLoki_CovarianceMap}
    \end{figure*}

    \begin{figure*}[t]
        \centering
        \includegraphics[width=0.9\linewidth]{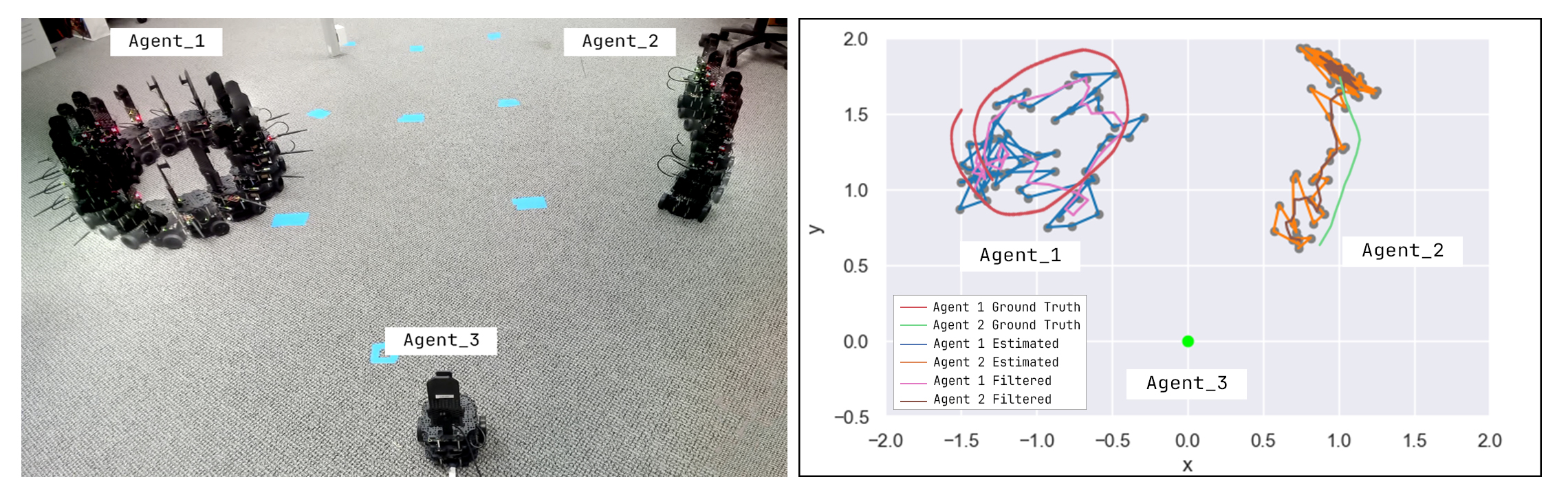}    
        \caption{Localization experiment with ReLoki. On the left we have the overlay of frames from the video captured from the motion of agents. Agent 1 is executing a circular motion and agent 2 is executing a curved foreword motion. On the right we have the output from ReLoki as seen by agent 3. We show both the raw estimation data and a data filtered using a low pass filter in the graph.}
        \label{fig:expm_localznoutput}
    \end{figure*}

    The covariance map is obtained using two factors. First we obtain the average error from the ground truth data~$\text{Cov}^e(q_m) := E[\hat{x}_{i, k}(q_m)^{\text{raw}} - x_{i, k}(q_m)]^2$ for all 50 readings taken at the relative range-pan-tilt $q_m$. The ground truth is obtained from the pan and tilt mechanism moving the ReLoki system to a specific relative bearing with respect to the source and the distance to that source set up during that experimental run. We also get the covariance (spread) of these 50 readings at the set relative position  . The final error measured is a very conservative measurement of the actual covariance and computed as shown in~\eqref{eqn:Covariance_ConservativeExt}. A lower value for this means a low cost in \eqref{eqn:objective}.


    \begin{align}
        \label{eqn:Covariance_ConservativeExt}
        \text{Cov}(q_m) = \det|\text{diag}(\text{Cov}^e(q_m)) + \text{Cov}^\sigma(q_m)|.
    \end{align}
    
    We show the measured covariance map in figure~\ref{fig:ReLoki_CovarianceMap}. Looking at the covariance maps, we can see that there are very high errors of more than $100\unit{\degree}$ at elevation angles $-90\unit{\degree}$, $-75\unit{\degree}$ and $-60\unit{\degree}$. This can be attributed to the electronics for ReLoki being in the path of the incoming signal at these angles, thereby skewing the estimates. We also see higher errors at elevation angle of $90$ degrees with an error of upto $80\unit{\degree}$, which may be due to antenna characteristics of the chosen antenna which has very little illumination from the source at this relative angle. Additionally, we  see that at elevation angles $-15\unit{\degree}$ - $-45 \unit{\degree}$ and close to azimuth of $-30$ and $+60$ there are higher errors in the range of upto $45 \unit{\degree}$ deviation from actual values. We attribute this to the ground plane of the protruding antenna $A_4$ interfering with Phase calculation in antenna $A_0$ and $A_1$. In other regions, the error in azimuth and elevation angles are within $15\unit{\degree}$ on average and will provide decent localization performance. The range measurements during the TWR ranging phase comes to an accuracy of about $25\unit{cm}$ on average up to a range measurement of $7\unit{m}$. We were getting spotty reception with completing both TWR and AOA measurements at and above $8\unit{m}$.

    We additionally show the comparison of RTA Array to our previous work, which uses orthogonal antenna array, in figure~\ref{fig:ReLoki_CovarianceMap}. Here we can observe a significant reduction in errors for higher elevation angles as noted by the green regions in the figure. This was the main weak point of the orthogonal array. The lower elevation angles, the errors are low in both implementations and the difference is very small here except for a few conditions as mentioned above. We can conclude that in applications where the elevation angles between agents are restricted to lower values, the Orthogonal array provides a slightly better estimate in the operational regions of such application. However, for a full 3D localization system, the RTA array provides overall better performance than the orthogonal array over the entire operational region. 

    In this system, we have observed that each RPP between agents takes up to $46\unit{ms}$ for based on clock output observed between transfers. This included handshake, one data packet, TWR and AOA calculation, which is the smallest transaction possible. Hence, we can estimate at-most 20 transactions between agents per second, which decreases with length of data being transmitted. So for $N_s$ agents the max frequency for each agent will be $20/(N_s - 1)$.

    To show the actual performance of the system in real robotic systems, we set up a 3 robot system: Two agents (Agent 1 and 2) in motion and the other (Agent 3) static. Agent 1 is executing a circular motion and Agent 2 is executing a curved forward trajectory as shown in figure~\ref{fig:expm_localznoutput}. We then capture the relative position measured by agent 3 of the other two agents. Note that the estimating agent can also be moving but we leave it fixed to have more accurate ground-truth measurements. 

    We also show the localization data output of the experiment in figure~\ref{fig:expm_localznoutput}. Here we can crudely see the loop created by agent 1 and the curved trajectory by agent 2 reflected in the output. The output is however very noisy with many jumps in estimated positions. Here it will be prudent to investigate the effects of a filter tuned to the specific motion type of the mobile platform; which will be one of the areas of investigation going forward. Additionally, we can also use sensor fusion of IMU Data with data obtained from ReLoki to further improve the localization estimates. 
 
    \section{Conclusions}
    In this paper we propose a novel UWB based relative localization method called ReLoki that leverages Angle of Arrival information in tandem with traditional ranging measurements to estimate the 3D relative position to any other participating agent. Our method is validated on an experimental platform that uses a Regular Tetrahedral antenna array and looking at the performance characteristics, we have shown the system using the RTA Array does better in high elevation readings, thereby works better in situations where full 3D localization is required as compared to our previous implementation using an orthogonal array. However, these high elevation errors are still comparatively higher than the low elevation errors and we need more research into antenna characteristics (e.g., effects of omnidirectionality and polarization) and other algorithmic improvements (per antenna pair filtering, multiple concurrent phase measurements in AoA phase) to tackle that. This extension from our previous system shows the flexibility of the platform to use any antenna array consisting of 4 elements. Further, this system can be easily integrated with any existing robotic platform and can be deployed in any unknown environment as it does not require any external infrastructure. 

    One of the main application of ReLoki platform is infrastructure-free localization in indoor aerial systems like Lighter-than-Air (LTA) agents, where the low weight of the platform is a plus. However, in such applications, we need to also study the Non Line of Sight (NLOS) characteristics of this system, particularly when these agents obstruct Line of Sight(LOS) detection.
    
    \section*{Acknowledgments}
    
    This work was supported in part by the Department of the Navy, Office of Naval Research (ONR), under federal grant N00014-20-1-2507.

    \printbibliography

\end{document}

%% file: macros.tex
\newcommand{\robots}{\mathcal{R}_s}
\newcommand{\robotsnum}{\mathcal{N}_s}
\newcommand{\robotposown}{p_i}

\newcommand{\robotbodyframe}{\prescript{B_j}{O}R_j}

\newcommand{\relpos}{q_{i, j}}
\newcommand{\relposest}{\hat{q}_{i, j}}

\newcommand{\carrierwl}{\lambda_c}
\newcommand{\uwbfptime}{t_{i, j}^{n.\text{fp}}}
\newcommand{\uwbphcalc}[1]{\Phi_{i, j}^#1}
\newcommand{\uwbphdiff}{\Delta\Phi_{i, j}^{n, o}}
\newcommand{\uwbphdifflim}{\Delta \bar{\Phi}_{i, j}^{n, o}}
\newcommand{\uwbphcal}{\tilde{\Phi}_j^{n, o}}
\newcommand{\uwbincang}{\alpha_{i, j}^{n, o}}
\newcommand{\uwbincangname}[2]{\alpha_{i, j}^{#1, #2}}

\newcommand{\urange}{r_{i, j}}
\newcommand{\ubear}{\hat{\upsilon}_{i, j}}
\newcommand{\ubearraw}{\hat{\upsilon}_{i, j}^{-}}
\newcommand{\ux}{\hat{\upsilon}_{i, j}^x}
\newcommand{\uy}{\hat{\upsilon}_{i, j}^y}
\newcommand{\uz}{\hat{\upsilon}_{i, j}^z}
\newcommand{\uxm}{\hat{\upsilon}_{i, j}^{x-}}
\newcommand{\uym}{\hat{\upsilon}_{i, j}^{y-}}
\newcommand{\uzm}{\hat{\upsilon}_{i, j}^{z-}}
\newcommand{\um}{\hat{\upsilon}_{i, j}^m}

\newcommand{\umm}{\hat{\upsilon}_{i, j}^{m-}}
\newcommand{\us}{\hat{\upsilon}_{i, j}^{s-}}

%% file: resources/common.tex
\newtheorem{theorem}{Theorem}[section]

\newtheorem{problem}[theorem]{Problem}

\newcommand{\real}{{\mathbb{R}}}

\renewcommand{\tilde}{\widetilde}

\renewcommand{\epsilon}{\varepsilon}

\renewcommand{\hat}{\widehat}

\newcommand{\oprocendsymbol}{\hbox{$\bullet$}}
\newcommand{\oprocend}{\relax\ifmmode\else\unskip\hfill\fi\oprocendsymbol}

\renewcommand{\theenumi}{(\roman{enumi})}

\newcommand{\comment}[1]{}

\definecolor{ashgrey}{rgb}{0.7, 0.75, 0.71}